\documentclass[11pt,a4paper]{article}
\usepackage[hyperref]{acl2020}
\usepackage{times}
\usepackage{latexsym}

\usepackage{todonotes}

\usepackage{microtype}
\usepackage{amsmath}
\usepackage{graphicx}
\aclfinalcopy

\usepackage{booktabs} 

\usepackage{multirow}
\usepackage{adjustbox}

\usepackage[format=hang, justification=RaggedRight]{subcaption}

\title{Data Science Kitchen at GermEval 2021: A Fine Selection of Hand-Picked Features, Delivered Fresh from the Oven}
\author{Niclas Hildebrandt, Benedikt Boenninghoff, Dennis Orth, Christopher Schymura \\
  Data Science Kitchen \\
  \texttt{\{firstname.lastname\}@data-science-kitchen.de} 
  }
\date{}

\begin{document}
\maketitle
\begin{abstract}
This paper presents the contribution\footnote{Code available on GitHub: \url{https://github.com/data-science-kitchen/germ-eval-2021}} of the Data Science Kitchen at GermEval 2021 shared task on the identification of toxic, engaging, and fact-claiming comments. The task aims at extending the identification of offensive language, by including additional subtasks that identify comments which should be prioritized for fact-checking by moderators and community managers. Our contribution focuses on a feature-engineering approach with a conventional classification backend. We combine semantic and writing style embeddings derived from pre-trained deep neural networks with additional numerical features, specifically designed for this task. Classifier ensembles are used to derive predictions for each subtask via a majority voting scheme. Our best submission achieved macro-averaged F1-scores of 66.8\%,\,69.9\% and 72.5\% for the identification of toxic, engaging, and fact-claiming comments.

\end{abstract}

\setlength{\abovedisplayskip}{8.pt}
\setlength{\belowdisplayskip}{10.pt}
\setlength{\abovedisplayshortskip}{8.pt}
\setlength{\belowdisplayshortskip}{10.pt}

\section{Introduction}
\label{sec:introduction}
In the early years after establishing social media platforms, setting up online discussion forums and installing comment areas on newspapers' websites, a door into this new digital world has been opened, allowing people to interconnect all over the world. Various communication platforms and social networks enabled new ways of sharing information with followers, exchanging opinions between politically interested people, and encouraging debates with the readers.
Unfortunately, recent trends revealed the ugly face and adverse effects of these platforms when an increasing number of users make improper, illegal, or abusive use of such digital services~\cite{10.1145/3292522.3326034}.

Nowadays, social media platforms are notorious for spreading toxic comments, in which the writers justify violence and discrimination against a person or groups of persons~\cite{Munn2020}. Additionally, a second steadily growing trend is producing and sharing fake news or misinformation, seeking to dominate current discussions, and frame public debates~\cite{Mahid2018}.

Both hate speech, fake news and their impact have become very prominent in recent years. However, the tremendous amount of shared and distributed toxic messages on social media platforms make it utterly infeasible to identify and tag or delete poisonous comments manually. The GermEval 2021 shared task tries to encounter this negative trend and motivates participants to work on automated solutions towards safer and more reliable digital rooms of interaction~\cite{germeval2021overview}.

Therefore, the organizers of the task increased the difficulty of the competition by expanding the focus not only on the identification of toxic messages in online discussions but also on distinguishing between engaging and fact-claiming comments. The first task is similar to the GermEval tasks in 2018~\cite{Wiegand2018} and 2019~\cite{stru-etal-2019-overview} and deals with identifying toxic comments, including offensive, hateful and vulgar language or ruthless cynism. As novel subtasks, the participants are also invited to identify two additional categories of comments: The second category defines engaging comments, which are annotated as highly relevant contributions by the moderators. The third category concentrates on finding fact-claiming comments that should be considered for a manual fact-check with a higher priority.

\section{Task and Data Description}
\label{sec:task_and_data}
Each subtask of GermEval 2021 is defined as a binary classification problem and all tasks share the same training and test data. The set of training data consists of 3,244 Facebook comments from a German news broadcast page. The anonymized comments were posted in the time span from February to July 2019 and were labeled by trained annotators. Binary labels were provided for each of the three categories. The test data is also extracted from Facebook discussions and include 944 comments. However, these comments had a different discussion topic than the training data. Precision, recall, and macro-averaged F1-score were defined as the relevant evaluation metrics.

\subsection{Subtask 1: Toxic Comment Classification}
Toxic comments are characterized by their offensive and hateful language, intended to blame other people or groups. For social media and content providers, it is important to detect such comments in a highly automated and scalable way. An example of a toxic comment from the training data of the GermEval shared task is: \textit{``Na, welchem tech riesen hat er seine Eier verkauft..?''}. However, some of the comments which have been labeled as toxic can be quite hard to detect. Examples of such cases are: \textit{``@USER eididei sieh mal an''} or \textit{``ein schöner VW Golf Diesel..''}. Difficulties occur due to irony, subtle overtones and missing contextual information.

\subsection{Subtask 2: Engaging Comment Classification}
Engaging comments encourage other users to join the discussion, express their opinions and share ideas regarding the topic. They are characterised by being rational, respectful, and reciprocal and hence can foster a constructive and fruitful discussion. The comment \textit{``Wie wär’s mit einer Kostenteilung. Schließlich haben beide Parteien (Verkäufer und Käufer) etwas von der Tätigkeit des Maklers. Gilt gleichermassen für Vermietungen. Die Kosten werden so oder soweit verrechnet, eine Kostenreduktion ist somit nicht zu erwarten.''} is an example of an engaging comment from the training data.

\subsection{Subtask 3: Fact-Claiming Comment Classification}
If a platform provider has to prevent the spread of fake news and misinformation, there is the demand of automatically identifying fact-claiming comments to assess their truthfulness. An example of a fact-claiming comment from the training data is the comment \textit{``Dummerweise haben wir in der EU und in der USA einen viel höheren CO2 Fußabdruck als z.B. die Afrikaner oder Inder.''}.

\section{System Overview}
\label{sec:system_overview}
\begin{figure*}[t]
\centering
\includegraphics[width=\textwidth]{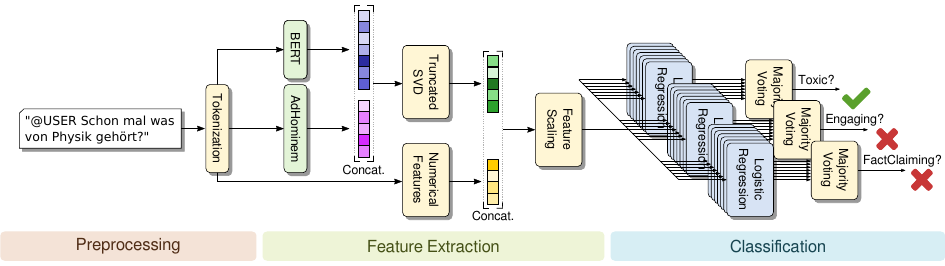}
\vspace*{-.6cm}
\caption{General architecture of the proposed framework to detect toxic, engaging and fact-claiming comments. Yellow boxes denote non-trainable, computational functions and transformations, green boxes represent pretrained models utilized for feature extraction. Trainable models whose parameters are optimized using the challenge dataset are shown in blue.}
\label{fig:architecture}
\vspace*{-0.25cm}
\end{figure*}

\noindent The general system architecture is shown in Fig.~\ref{fig:architecture}. As the number of samples in the training dataset provided for GermEval 2021 is rather small, our proposed framework focuses on suitable feature engineering with a conventional classification backend. These features and further implementation details of our system are described in the following.

\subsection{Preprocessing}
\label{subsec:preprocessing}
Raw input text is preprocessed in three different processing streams that are handled in parallel. The first stream utilizes the tokenizer of a German BERT model~\cite{Chan2020} and crops the corresponding input text at a maximum length of 512 tokens. The second stream uses the SoMaJo tokenizer~\cite{Proisl2016} for German language and the third stream passes the raw text to the subsequent feature extraction stage without any preprocessing.

\subsection{Feature Extraction}
\label{subsec:feature_extraction}
The feature extraction stage focuses on embedding-based features, as well as manually selected, numerical feature representations. Specific feature types are computed using one of the three preprocessing streams described in Sec.~\ref{subsec:preprocessing}. This specific feature extraction setup was chosen to efficiently combine embedding representations that capture linguistic properties with ``hand-crafted'' features specifically designed for the GermEval 2021 tasks.

\subsubsection{Semantic Embeddings}
\label{subsec:semantic_embeddings}
The first kind of embeddings used in our framework are document embeddings derived from a pre-trained German BERT model~\cite{Chan2020}. Specifically, we used the \texttt{bert-base-german-cased} implementation from Huggingface\footnote{\url{https://huggingface.co/dbmdz/bert-base-german-cased}}. This model was trained on a German Wikipedia dump, the OpenLegalData dump~\cite{Ostendorff2020} and news articles. Average pooling was used to compute 768-dimensional document-level embeddings from the BERT model output.

\subsubsection{Writing Style Embeddings}
\label{subsec:writing_style_embeddings}
Besides semantic document embeddings, we additionally experimented with neural stylometric embeddings that have been automatically extracted from the comments. More precisely, we used an extended framework of \textsc{AdHominem}~\cite{boenninghoff:2019b} which outperformed all other systems that participated in the PAN 2020 and 2021 authorship verification tasks~\cite{boenninghoff:2021}.

The overall framework consists of three components: In a first step, we perform neural feature extraction and deep metric learning (DML) to encode the writing style characteristics of a pair of raw documents into a pair of fixed-length representations, which is realized in the form of a Siamese network. 
Inspired by~\cite{HochSchm97}, the Siamese network consists of a hierarchical LSTM-based topology.
Next, the obtained representations are fed into a Bayes factor scoring (BFS) layer to compute the posterior probability for this trial. 
The idea of this second component is to take into account both, the similarity between the questioned documents and the typicality w.r.t. the relevant population represented by the training data.
The third component is given by an uncertainty adaptation layer (UAL) aiming to correct possible misclassifications and to return corrected and calibrated posteriors. More details can be found in~\cite{boenninghoff:2021}.

To train the model, we prepared a large dataset of Zeit-Online forum comments\footnote{\url{www.zeit.de}}.
Altogether, we collected 9,812,924 comments written by 204,779 authors. Afterwards we split the dataset into training and validation sets. We took 10\% of the authors to build the validation set and removed all comments with less than $60$ tokens. Due to the fact that the provided dataset of the shared task also contains concise comments, we decided to leave all short comments in the training set. As a result, the datasets are disjoint w.r.t. the authors, i.e., all authors in the validation set have been removed from the training set.
During training, we perform data augmentation by resampling new same-author and different-authors pairs in each epoch. Contrary, the pairs of the validation set are sampled once and then kept fixed. Since some authors contribute with hundreds of comments, we limited their influence by sampling not more than 20 comments per author. 
In summary, the training set contains approximately 234,500 same-author and 244,200 different-authors pairs in each epoch, where, on average, each comment consists of $75.90 \pm 68.07$ tokens.
The validation set contains 15,125 same-author and 18,740 different-authors pairs, where, on average, each comment consists of $126.51 \pm 65.31$ tokens. Hence, both datasets are nearly balanced. 

\begin{table}[t]
\centering
\caption{Results for PAN 2021 evaluation metrics.}
\vspace*{-0.1cm}
\label{tab:results:adhominem}
\resizebox{1.0\columnwidth}{!}{
    \begin{tabular}{|c c | c |c |c |c |c |c|}
    \hline
   \multicolumn{2}{|c|}{\multirow{2}{*}{\textbf{Model}}}
    & \multicolumn{6}{c|}{\textbf{PAN 2021 Evaluation Metrics}}
       \\ \cline{3-8}
    & &{AUC}           &{c@1/acc}       &{f\_05\_u}    &{F1}  &{Brier}      &{Overall}
    \\ \hline
\multicolumn{2}{|c|}{DML}
    &$87.4$  
    &$79.3$  
    &$81.7$  
    &$81.0$ 
    &$85.1$ 
    &$82.9$
    \\
\multicolumn{2}{|c|}{BFS}
    &$87.4$  
    &$79.5$  
    &$80.5$  
    &$82.0$ 
    &$85.5$ 
    &$83.0$ 
        \\ 
\multicolumn{2}{|c|}{UAL} 
    &$87.6$  
    &$79.5$  
    &$81.6$  
    &$81.4$ 
    &$85.6$ 
    &$83.2$ 
    \\ \hline
    \end{tabular}
}
\vspace*{-0.4cm}
\end{table}

\begin{table*}[t]
\centering
\caption{Overview of all features utilized in this work that are not based on embeddings.}
\vspace*{-0.2cm}
\label{tab:numerical_features}
\resizebox{1.0\textwidth}{!}{
    \begin{tabular}{|l|c|l|}
    \hline
    \textbf{Feature name} & \textbf{Dim.} & \textbf{Description} \\
    \hline
    \texttt{NumCharacters} & 1 & Total number of characters, including white spaces. \\
    \texttt{NumTokens} & 1 & Total number of tokens, after splitting at white spaces. \\
    \texttt{AverageTokenLength} & 1 & Average number of characters in all tokens. \\
    \texttt{TokenLengthStd} & 1 & Standard deviation of the number of characters in all tokens. \\
    \texttt{StopwordRatio} & 1 & Number of stop words divided by the number of tokens. \\
    \texttt{ExclamationMarkRatio} & 1 & Number of exclamation marks divided by the number of characters. \\
    \texttt{NumReferences} & 1 & Number of hyperlinks in the comment. \\
    \texttt{NumMediumAdressed} & 1 & Number of \texttt{@MEDIUM} mentions in the comment. \\
    \texttt{NumUserAdressed} & 1 & Number of \texttt{@USER} mentions in the comment. \\
    \texttt{AverageEmojiRepetition} & 1 & Average repetition number of emojis used in the comment. \\
    \texttt{SpellingMistakes} & 17 & Number of specific grammar and spelling mistakes, cf. Sec.~\ref{subsec:numerical_features}. \\
    \texttt{SentimentBERT} & 3 & Sentiment scores of a pre-trained BERT model~\cite{Guhr2020}. \\
    \hline
    \end{tabular}
}
\vspace*{-0.2cm}
\end{table*}

%\begin{table*}[t]
%\centering
%\caption{Results for PAN 2021 evaluation and calibration metrics.}
%\label{tab:results:adhominem}
%\resizebox{0.7\textwidth}{!}{
%    \begin{tabular}{|c c | c |c |c |c |c |c|c|c|c|}
%    \hline
%   \multicolumn{2}{|c|}{\multirow{2}{*}{\textbf{Model}}}
%    & \multicolumn{6}{c|}{\textbf{PAN 2021 Evaluation Metrics}}
%    & \multicolumn{3}{c|}{\textbf{Calibration Metrics}}
%       \\ \cline{3-11}
%    & &\texttt{AUC}           &\texttt{c@1/acc}       &\texttt{f\_05\_u}    &\texttt{F1}  &\texttt{Brier}      &\texttt{overall} 
%    &\texttt{conf}   
%    &\texttt{ECE}           
%    &\texttt{MCE}
%    \\ \hline
%\multicolumn{2}{|c|}{DML}
%    &$87.4$  
%    &$79.3$  
%    &$81.7$  
%    &$81.0$ 
%    &$85.1$ 
%    &$82.9$
%    &$72.4$
%    &$6.8$ 
%    &$9.6$
%    \\
%\multicolumn{2}{|c|}{BFS}
%    &$87.4$  
%    &$79.5$  
%    &$80.5$  
%    &$82.0$ 
%    &$85.5$ 
%    &$83.0$ 
%    &$75.3$
%    &$4.2$ 
%    &$5.9$
%        \\ 
%\multicolumn{2}{|c|}{UAL} 
%    &$87.6$  
%    &$79.5$  
%    &$81.6$  
%    &$81.4$ 
%    &$85.6$ 
%    &$83.2$ 
%    &$76.0$
%    &$3.4$ 
%    &$4.2$
%    \\ \hline
%    \end{tabular}
%}
%\end{table*}

We choose the PAN 2021 evaluation metrics to evaluate the performance as described in~\cite{kestemont:2021}. %In addiiton, we provide the averaged confidence score (\texttt{conf}), expected calibration error (\texttt{ECE}) and maximum calibration error (\texttt{MCE})~\cite{pmlr-v70-guo17a} to capture the calibration capacity.
Table.~\ref{tab:results:adhominem} summarizes the results, where all three system components are evaluated separately. 
It can be seen that we achieved overall scores between $82.9$ and $83.2$ for the components, which is mainly supported by higher values for the AUC and Brier scores. Comparing the c@1, F1 or f\_05\_u metrics, we generally obtained error rates of approximately $20\%$ on this challenging dataset for a fixed threshold. 
%Similar to the results in~\cite{boenninghoff:2021}, the \texttt{ECE} and \texttt{MCE} show significant improvements for the \texttt{BFS} and \texttt{UAL} components, while the overall score slightly increases from $82.9$ to $83.2$.
After training, one part of the neural feature extraction component within the Siamese network is then used to extract the 100-dimensional writing style embeddings for the shared task data.

\subsubsection{Additional Numerical Features}
\label{subsec:numerical_features}
In addition to the semantic and writing style embeddings, we integrated a set of specifically designed numerical features into our framework. An overview of these features, their dimensionality and corresponding descriptions is given in Tab.~\ref{tab:numerical_features}. We applied the natural logarithm to all strictly-positive numerical features.

The first group of features, \texttt{NumCharacters}, \texttt{NumTokens}, \texttt{AverageTokenLength} and \texttt{TokenLengthStd}, were chosen to reflect general structural properties of the comments in the dataset. In addition, we use the \texttt{StopwordRatio} and \texttt{ExclamationMarkRatio} features to explicitly reflect task-related semantic properties in the dataset. These task-specific features are accompanied by additional count-based features \texttt{NumMediumAdressed}, \texttt{NumUserAddressed}, \texttt{NumReferences} and \texttt{AverageEmojiRepetition}. We also included the scores (corresponding to the classes ``positive'', ``neutral'' and ``negative'') of a BERT model for sentiment classification trained on 1,834 million German-language samples derived from various sources~\cite{Guhr2020} as a dedicated \texttt{SentimentBERT} feature.

Lastly, we included an 17-dimensional feature denoted as \texttt{SpellingMistakes} into our set of additional features. This feature represents spelling and grammar mistakes from 17 different categories. We used a Python wrapper from the open-source grammar checker \texttt{LanguageTool}\footnote{\url{https://languagetool.org/}} to derive this feature. In particular, the following classes of mistakes were considered: \textit{Typography}, \textit{punctuation}, \textit{grammar}, \textit{upper/lowercase}, \textit{support in punctuation}, \textit{colloquialism}, \textit{compounding}, \textit{confused words}, \textit{redundancy}, \textit{typos}, \textit{style}, \textit{proper nouns}, \textit{idioms}, \textit{recommended spelling}, \textit{miscellaneous}, \textit{double punctuation}, \textit{double exclamation mark}. For every category, we counted the number of mistakes and divided them by the number of tokens in the respective comment.

\subsection{Classification Pipeline}
\label{subsec:classification_pipeline}
The classification pipeline used in this work is depicted in Fig.~\ref{fig:architecture}. The semantic and writing style embedding features described in Secs.~\ref{subsec:semantic_embeddings} and~\ref{subsec:writing_style_embeddings} are concatenated, yielding a 868-dimensional joint embedding vector. A truncated singular-value decomposition (SVD)~\cite{Halko2011} is applied on this vector to reduce its dimensionality for subsequent processing. The number of dimensions kept is treated as a hyperparamter during training, cf. Sec.~\ref{sec:evaluation}. The reduced joint embedding vector is then concatenated with the 28-dimensional vector of additional numerical features. The resulting vector is standardized to zero-mean and unit variance and serves as input to the classification stage.

We use Logistic Regression~\cite{Berkson1944, Haggstrom1983} and Support Vector Machines (SVMs)~\cite{Boser1992} with radial basis function (RBF) kernel as base classifiers within individual ensembles. One ensemble of binary classifiers is utilized for each subtask. Each classifier in the ensembles is trained using a subset of the provided training data via a cross-validation setup, cf. Sec.~\ref{sec:evaluation}. A hard majority-voting scheme is used in each ensemble to obtain the predicted labels.

% The difference between the first two submissions and submission 3 is that an extended voting ensemble is used with half of the models being SVMs with hyperparameters which have been tuned individually for each task and that the Adhominem features are not included.
\begin{figure*}[t]
\centering
\begin{subfigure}[t]{0.3\textwidth}
\centering
\includegraphics[width=0.99\textwidth]{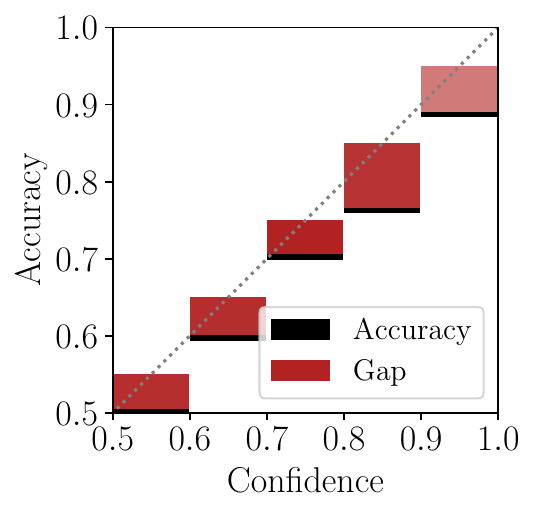}
\vspace*{-0.7cm}
\caption{Subtask 1}
\end{subfigure}
\begin{subfigure}[t]{0.3\textwidth}
\centering
\includegraphics[width=0.99\textwidth]{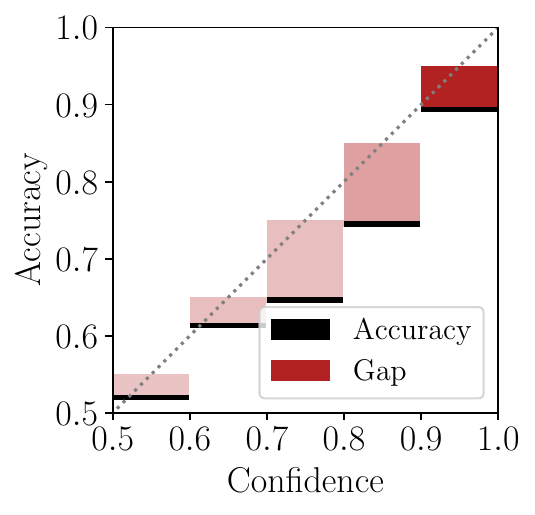}
\vspace*{-0.7cm}
\caption{Subtask 2}
\end{subfigure}
\begin{subfigure}[t]{0.3\textwidth}
\centering
\includegraphics[width=0.99\textwidth]{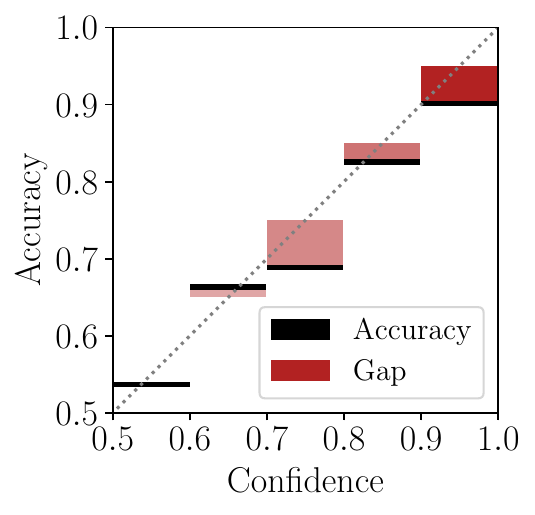}
\vspace*{-0.7cm}
\caption{Subtask 3}
\end{subfigure}
\vspace*{-0.2cm}
\caption{Reliability diagrams of our first submission for all three subtasks (see Section~\ref{subsec:evaluation_metrics}). The red bars are drawn darker for bins with a higher number of samples.} 
\label{fig:reliability_diagram}
\vspace*{-0.2cm}
\end{figure*}
% \footnote{\url{https://github.com/hollance/reliability-diagrams}}

\section{Evaluation}
\label{sec:evaluation}
Our framework is trained using a specific cross-validation and hyperparameter tuning scheme, which is described in the following.

\subsection{Evaluation Metrics}
\label{subsec:evaluation_metrics}
Precision, recall and macro average F1-score are used for model evaluation~\cite{opitz2021macro} since they represents the evaluation metrics of the GermEval 2021 shared task. Additionally, we assess the calibration properties of our model by determining the expected calibration error (ECE) as well as the maximum calibration error (MCE), where the confidence interval is discretized into a fixed number of $M$ bins~\cite{Naeini2015calibration}. The ECE is then computed as the weighted macro-averaged absolute error between confidence and accuracy of all bins, %
\begin{equation}
    \mathrm{ECE} = \sum_{m=1}^{M}\frac{|B_m|}{N} \left| \mathrm{acc}(B_m) - \mathrm{conf}(B_m)\right|,
    \label{eq:ECE}
\end{equation}
where $N$ is the total number of samples and $\mathrm{acc}(B_m) - \mathrm{conf}(B_m)$ is the difference between the actual accuracy and classifier confidence within a fixed-size bin $B_m$ in the confidence interval.
Note that all confidence values lie within the interval $[0.5,1]$,since we are dealing with binary classification tasks. Hence, to obtain confidence scores, the output predictions $p$ are  transformed  w.r.t.  to  the  estimated  subtask label, showing $\mathrm{conf}=p$ if the $\mathrm{acc} >= 0.5$ and $\mathrm{conf}=1-p$ if $\mathrm{acc} < 0.5$.
The MCE returns the maximum absolute error, given as
\begin{equation}
    \mathrm{MCE} = \max_{m \in {1,..., M}} \left| \mathrm{acc}(B_m) - \mathrm{conf}(B_m)\right|.
    \label{eq:MCE}
\end{equation}
We further display the reliability diagrams in Fig.~\ref{fig:reliability_diagram} which will be discussed in Section~\ref{sec:results_and_discussion}.

%

%To measure the deviation between confidence and accuracy for the worst-case scenario the Maximum Calibration Error (MCE) is defined as
%
%Visually speaking: The MCE is the largest calibration gap (red bar) and ECE is the weighted average of all gaps in figure~\ref{fig:reliability_diagram} as derived in
%~\cite{guo2017calibration}.

%\begin{figure}[h]
%\centering
%\includegraphics[width=0.5\textwidth]{figs/density_plot.pdf}
%\caption{Kernel density plot.} 
%\end{figure}
%\todo{Explanation}

\subsection{Experimental Setup}

\begin{table*}[t]
\centering
\caption{Final submission results on the test set including the calibration metrics for the first submission.}
\vspace*{-0.2cm}
\label{tab:submission_results}
\resizebox{1.0\textwidth}{!}{
    \begin{tabular}{|cc || c|c|c|c|c 
                        || c|c|c|c|c   
                        || c|c|c|c|c|
                        }
    \hline
   \multicolumn{2}{|c||}{\multirow{2}{*}{\textbf{Run}}}
    & \multicolumn{5}{c||}{\textbf{Subtask 1}}
    & \multicolumn{5}{c||}{\textbf{Subtask 2}}
    & \multicolumn{5}{c|}{\textbf{Subtask 3}}
   \\ \cline{3-17}
    & &\texttt{P}           
    &\texttt{R}       
    &\texttt{F1}    
    &\texttt{ECE}           
    &\texttt{MCE}
    &\texttt{P}  
    &\texttt{R}      
    &\texttt{F1} 
    &\texttt{ECE}           
    &\texttt{MCE}
    &\texttt{P}  
    &\texttt{R}      
    &\texttt{F1} 
    &\texttt{ECE}           
    &\texttt{MCE}
    \\ \hline\hline
\multicolumn{2}{|c||}{Submission 1}
    &$65.95$  
    &$63.67$  
    &$64.79$  
    &$5.5$
    &$~~8.0$
    &$69.70$  
    &$67.78$  
    &$68.72$  
    &$6.9$
    &$10.4$
    &$73.25$  
    &$71.44$  
    &$72.34$  
    &$3.5$
    &$6.0$
    \\ \hline
\multicolumn{2}{|c||}{Submission 2}
    &$64.89$  
    &$62.71$  
    &$63.78$  
    &$-$
    &$-$
    &$69.26$  
    &$67.43$  
    &$68.33$  
    &$-$
    &$-$
    &$73.39$  
    &$71.52$  
    &$72.44$  
    &$-$
    &$-$
    \\ \hline
\multicolumn{2}{|c||}{Submission 3}
    &$66.98$  
    &$66.73$  
    &$66.85$  
    &$-$
    &$-$
    &$71.71$  
    &$68.34$  
    &$69.98$  
    &$-$
    &$-$
    &$73.03$  
    &$72.08$  
    &$72.55$  
    &$-$
    &$-$
    \\ \hline
    \end{tabular}
}
\end{table*}

Our experimental setup involves dedicated model selection and hyperparameter tuning. The training set performance is evaluated in a stratified $K$-Fold cross validation setup preserving the class label distribution among all folds. One of the $K$ folds is used as validation set. We utilized a 7-fold cross-validation scheme and computed the evaluation metrics described in Sec.~\ref{subsec:evaluation_metrics} on the validation set of each fold.

For submission one and two there are 7 logistic regression models for each subtask trained on different folds and stacked together in a voting ensemble returning the prediction of the majority. On each fold the L2-regularisation strength \texttt{C} and the number of features coming from the SVD dimension reduction are tuned with respect to the macro averaged F1 score over all subtasks. This means that hyperparameters may be slightly different from fold to fold but all three models trained on the same fold get the same hyperparameters -- regardless the classification task. 

Submission three uses a similar approach but the logistic regression models are replaced by SVMs having the same  fold-wise hyperparameter tuning as mentioned above. In addition, task-wise tuned SVMs are added to the ensemble. Doubling the number of models and including a higher level of customisation to the task. The task-wise tuning includes optimisation of \texttt{kernel}, L2 regularisation strength \texttt{C}, \texttt{class weight}(whether or not to weight \texttt{C} with the class label distribution) and the kernel coefficient \texttt{gamma} as defined in the sklearn library~\cite{scikit-learn}.

Hyperparameter tuning is performed with Bayesian optimisation using the Optuna library~\cite{akiba2019optuna}. The macro average F1-score is chosen as optimisation target and the best hyperparameters among 100 trails are used in the ensemble.

\begin{figure*}[t]
\centering
\begin{subfigure}[t]{0.28\textwidth}
\centering
\includegraphics[width=0.99\textwidth]{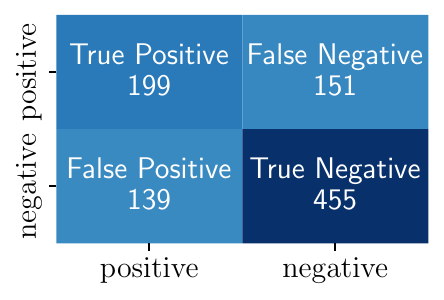}
\vspace*{-0.7cm}
\caption{Subtask 1}
\end{subfigure}
\begin{subfigure}[t]{0.28\textwidth}
\centering
\includegraphics[width=0.99\textwidth]{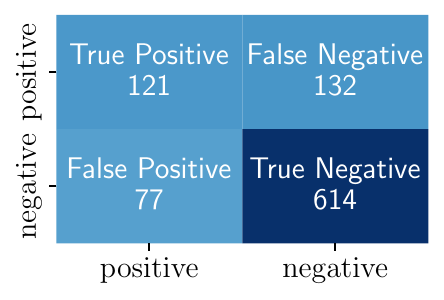}
\vspace*{-0.7cm}
\caption{Subtask 2}
\end{subfigure}
\begin{subfigure}[t]{0.28\textwidth}
\centering
\includegraphics[width=0.99\textwidth]{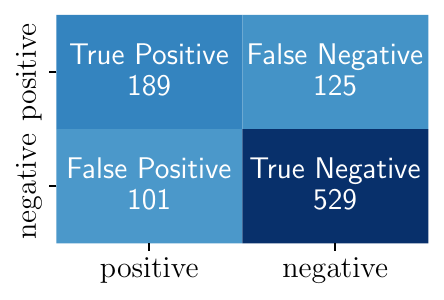}
\vspace*{-0.7cm}
\caption{Subtask 3}
\end{subfigure}
\vspace*{-0.3cm}
\caption{Confusion matrices for submission 3.} 
\label{fig:confusion_matrix_submission_3_small}
\end{figure*}

\section{Results and Discussion}
\label{sec:results_and_discussion}

The final submission results are provided in Table~\ref{tab:submission_results}. 
Unexpectedly, the identification of toxic comments turns out to be the most challenging subtask while the detection of fact-claiming comments achieved the highest F1-score.
This confirms our observations during hyperparameter tuning. 
For instance, the F1-score for the third submission after cross validation are given by $66.31 \pm 1.76$, $75.12 \pm 2.07$ and $74.68 \pm 2.67$ for subtasks 1-3, respectively.
A comparison of our cross validation performance with the results on the test set shows two interesting findings: On the one side, we obtained very robust results of subtasks 1 and 3. On the other side, subtask 2 struggles with over-fitting effects.

In addition, Fig.~\ref{fig:confusion_matrix_submission_3_small} displays the confusion matrices of our third submission (representative for all submissions).
It can be seen for all subtasks that the ratio of wrongly classified positively labeled samples is significantly larger than for negatively labeled samples. This behavior is supported by the reliabilty diagrams\footnote{\url{https://github.com/hollance/reliability-diagrams}} in Fig.~\ref{fig:reliability_diagram}, where our submission delivers \textit{over-confident} scores (i.e. conf $>$ acc) in nearly all bins.
As a results, the higher proportion of wrongly classified comments for positively labeled comments leads to a lower performance in terms of the F1-score.

%It can be seen that false positives and false negatives are nearly equally distributed for subtast 1 and 3. In the case of "Engaging" comment identification the model tends to have more false negative predictions than false positive predictions. Hence, classification predictions for the category "Engaging" tend to have a better accuracy in detecting engaging comments than detecting that it is not an engaging comment.

\begin{figure}[t]
\centering
\includegraphics[width=0.5\textwidth]{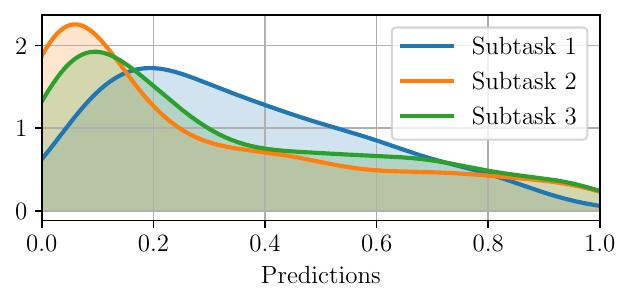}
\vspace*{-0.8cm}
\caption{Gaussian kernel density estimates for the distributions of the first submission (bandwidth$=0.08$).} 
\label{fig:kernel_density}
\vspace*{-0.1cm}
\end{figure}

Finally, we visualize an estimated probability density function  of the first submission using a non-parametric Gaussian kernel density estimator\footnote{\url{https://scikit-learn.org/stable/modules/generated/sklearn.neighbors.KernelDensity.html}} in Fig.~\ref{fig:kernel_density}.
Ideally, we would expect a bimodal probability density function. However, the plot shows that the system clearly tends towards self-confident predictions close to zero. But in regions closer to one, the systems behave more hesitant. This effect can be explained by the imbalanced distribution of the class labels. 

\section{Conclusions}
\label{sec:conclusions}
Within this contribution to the shared task of the GermEval 2021 we have developed a modular feature extraction scheme which incorporates semantic and writing style embeddings as well as task specific numerical features. Less complex algorithms like logistic regression models and SVMs converge converge faster than complex models like deep neural networks and therefore need less training data. The combination with automated hyperparameter tuning and dimension reduction as well as the final agglomeration of multiple models in voting ensembles allow to achieve an macro-averaged F1-scores of 66.8\%,\,69.9\% and 72.5\% for the identification of toxic, engaging, and fact-claiming comments.

%\clearpage
\bibliography{acl2020}
\bibliographystyle{acl_natbib}

\end{document}